\pdfoutput=1

\documentclass[11pt]{article}

\usepackage[final]{acl}

\usepackage{times}
\usepackage{latexsym}
\usepackage{float}
\usepackage[T1]{fontenc}

\usepackage[utf8]{inputenc}
\usepackage{microtype}

\usepackage{inconsolata}
\usepackage{booktabs}
\usepackage{multirow}
\usepackage{graphicx}
\usepackage{todonotes}
\usepackage{paralist}

\title{LEVOS: Leveraging Vocabulary Overlap with Sanskrit to Generate Technical Lexicons in Indian Languages}

  \author{Karthika N J, Krishnakant Bhatt, Ganesh Ramakrishnan, and Preethi Jyothi \\
  Indian Institute of Technology Bombay, India \\
  \texttt{\{karthika, kkbhatt, ganesh, pjyothi\}@cse.iitb.ac.in} \\
  }

\begin{document}
\maketitle
\begin{abstract}
Translating technical terms into lexically similar, low-resource Indian languages remains a challenge due to limited parallel data and the complexity of linguistic structures. We propose a novel use-case of Sanskrit-based segments for linguistically informed translation of such terms, leveraging subword-level similarity and morphological alignment across related languages. Our approach uses character-level segmentation to identify meaningful subword units, facilitating more accurate and context-aware translation. To enable this, we utilize a Character-level Transformer model for Sanskrit Word Segmentation (CharSS), which addresses the complexities of sandhi and morpho-phonemic changes during segmentation. We observe consistent improvements in two experimental settings for technical term translation using Sanskrit-derived segments, averaging 8.46 and 6.79 chrF++ scores, respectively. Further, we conduct a post hoc human evaluation to verify the quality assessment of the translated technical terms using automated metrics. This work has important implications for the education field, especially in creating accessible, high-quality learning materials in Indian languages. By supporting the accurate and linguistically rooted translation of technical content, our approach facilitates inclusivity and aids in bridging the resource gap for learners in low-resource language communities.

\end{abstract}

\section{Introduction}
English is the most widely used language in academic books and as a medium of instruction worldwide. India has 22 official languages in addition to English. Over the years, works like \citep{hudelson1987role} have established the power of language in students' learning.
Aligned with these studies, the Indian Government has proposed several changes to the education system, among which multilingual knowledge dissemination assumes an important role. The proposal involves introducing regional languages at various levels of education. Carrying out this proposal leads to massive resource requirements like textbook translations and content creation. 
With the limited technical content availability in non-English languages, especially at the higher education level, translating technical terms from English to other languages is a challenging task that needs to be addressed.\\
\citet{maheshwari2024lexgen} presents the importance of domain-specific lexicon generation, especially catering to the technical domains, and its importance for translation tasks with low-resource languages as the target. \citet{kunchukuttan2016faster} shows the importance of subword segmentation and lexical similarity of languages in the translation task. Additionally, Sanskrit language is known to be a lexically rich, flexible, and well-structured language with the potential to create meaningful new words easily.  In this paper, we introduce a use case of Sanskrit-based sub-word level segmentation in word and phrase-level translation of academic/technical terminologies to leverage the large overlap of vocabulary among Indian languages. For the generation of technical terms in low-resource regional languages, we propose to utilize the high vocabulary overlap of Indo-Aryan and Dravidian languages with Sanskrit, thereby performing a lexically informed translation. 

Compound words are formed by combining two or more meaningful subwords. In Indian languages, compounds may be formed either through simple concatenation without boundary changes or by following sandhi rules, resulting in boundary modifications.
Decompounding a compound Sanskrit word involves segmenting it into smaller, meaningful lexical units. 
Existing methods used for the Sanskrit Word Segmentation (SWS)\footnote{We use the term segmentation for the task of splitting a compound word into its meaningful constituents.} task can be roughly classified into two categories: tackling the broader task of SWS and sandhi splitting-specific techniques. The former includes works like \cite{gerard2003lexicon, sriram-etal-2023-validation}, a lexicon-driven shallow parser. \citet{hellwig-nehrdich-2018-sanskrit} processes compound sandhi words at the character level using recurrent and convolutional neural networks. \citet{sandhan2022translist} presents TransLIST, integrating a module that appends additional latent information from SHR to the input sequence. It also employs a soft masked attention mechanism to prioritize relevant subword candidates and incorporates a path ranking algorithm to mitigate erroneous predictions.
Alternately, \citet{aralikatte2018sanskrit} proposes a dual-decoder approach where the first decoder identifies the location for the sandhi split (sandhivicchēda)\footnote{We follow ISO-15919 script to mention Roman translations of Indian language text for better readability.}, and the second decoder predicts the segmented output. Similarly, \citet{dave2021neural} applies an RNN encoder-decoder-based two-stage methodology to predict the location and final splits. \citet{nehrdich-etal-2024-one} presented a new language model pre-trained for Sanskrit and further fine-tuned and utilised the model for various downstream tasks including word segmentation, lemmatization and morphosyntactic tagging tasks. We use a similar architecture in our CharSS model, to generate the word-splits.



Our main contributions through the paper are:
\begin{asparaitem}
    \item We present the utilization of a character-based Transformer model for the segmentation of compound words (including sandhivicchēda) in Sanskrit (Section~\ref{sec:word-segment}).
    \item We propose a Sanskrit-based input augmentation method using relatively resource-rich Hindi translations to generate linguistically informed technical lexicons for lexically similar, low-resource languages (Section~\ref{sec:ttg}). 
    \item Through comprehensive experiments, we show the efficacy of our proposed methodologies. We test CharSS on three benchmark datasets for SWS. Similarly, we experiment with our technical term translation process for multiple low-resource languages, generating better-quality technical lexicons in the target languages (Section~\ref{sec:exp}).
\end{asparaitem}

\section{Methodology}
\subsection{Sanskrit Word Segmentation}
\label{sec:word-segment}
\begin{figure}[h]
    \centering
    \includegraphics[width=0.5\textwidth]{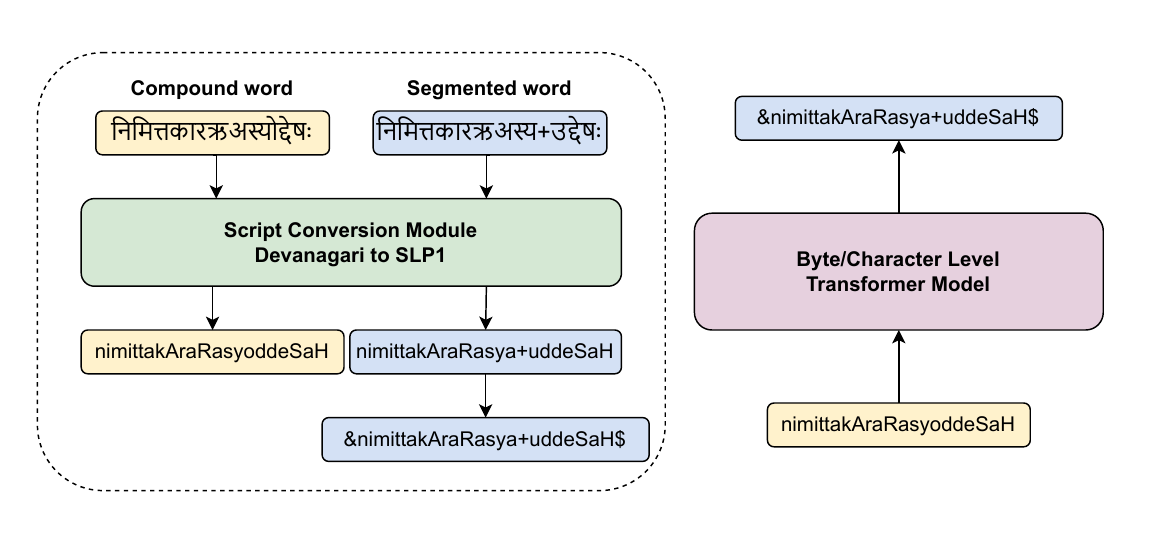}
    \caption{Illustration of the proposed methodology for SWS task.} 
    \label{fig:swsarch}
\end{figure}

Figure \ref{fig:swsarch} illustrates the proposed methodology for SWS. We formulate the task of sandhi splitting and Sanskrit Word Segmentation as a standalone sequence-to-sequence transformation problem. For this purpose, we propose to utilize a character-level Transformer model such as ByT5.

\paragraph{ByT5.} The ByT5 (Byte-Level Text-to-Text Transfer Transformer) model \cite{xue2022byt5} processes text as sequences of bytes, bypassing the need for language-specific tokenization. This approach enables it to handle diverse languages and scripts effectively, including rare words and complex orthographies. ByT5 is built on the T5 \cite{raffel2020exploring} framework. It poses all tasks as text-to-text problems, enhancing its versatility. ByT5 demonstrates strong performance on multilingual and code-mixed tasks, making it particularly suitable for low-resource languages and domain-specific vocabularies. The input to the model is a single Sanskrit word (unigram), and the output consists of the segmented sub-tokens of the word, which are concatenated using a "+" symbol to indicate the split. We prepend the target split with an "\&" symbol to denote the start and append a "\$" symbol to mark the end of the target split as shown in Figure \ref{fig:swsarch} to allow for precise delineation of morpheme boundaries. 

\subsection{Technical Term Translation}
\label{sec:ttg}
In this paper, we propose a linguistically informed method to translate technical terms in English to low-resource Indian languages. This process entails a crucial input augmentation phase prior to the modeling and training stages to enhance the input for model training. The raw dataset comprises technical terms for English and translation to Hindi. We prepare supplementary data for augmentation using the methodology described below.

\subsubsection*{Sanskrit-based augmented input }
\label{sec:sbad}
\begin{figure}
    \centering
    \includegraphics[width=0.5\textwidth]{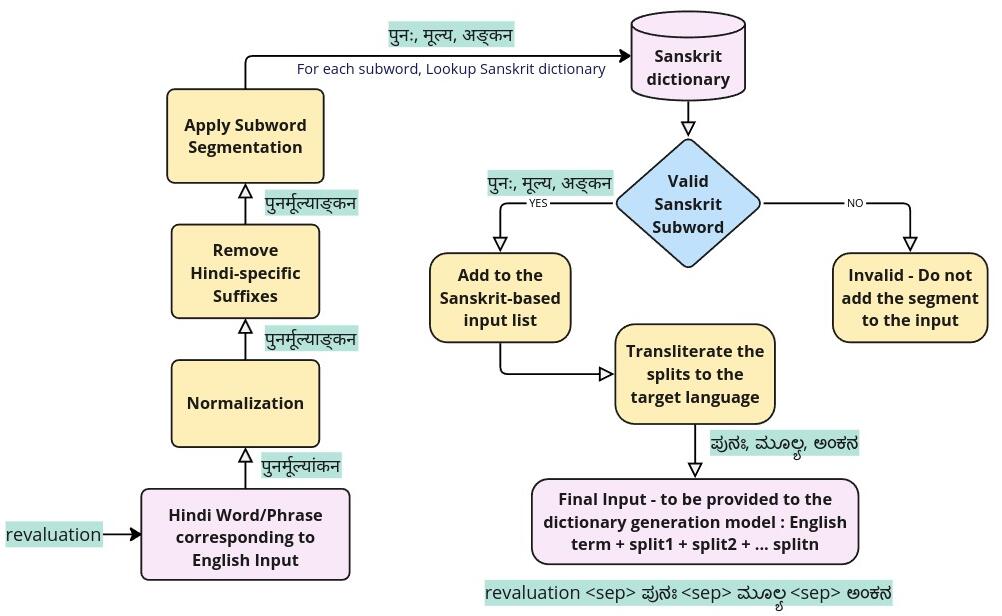}
    \caption{The process of generating Sanskrit-based augmented input for the English term '\textit{revaluation}', for translation model}
    \label{fig:san_input}
\end{figure}

There is a significant vocabulary overlap among Indian languages, especially with Sanskrit. In this work, we attempt to leverage this overlap by using available dictionaries in the resource-rich Hindi language to generate the corresponding terms in other Indian languages. Figure \ref{fig:san_input} shows the steps to obtain the proposed augmented input. For a given technical term, we first normalize the corresponding term in Hindi as explained in Appendix~\ref{sec:norm}. We then remove the Hindi-specific affixes from the words to get the lemma. Finally, we perform segmentation of the normalized lemma and pass them as additional input to the translation model to aid the generation of technical terms in low-resource Indian languages. 

\subsubsection*{Motivation to use Hindi data to generate Sanskrit-based segments:} 
There is a significant under-representation of digital resources for all other Indian languages compared to Hindi. Appendix \ref{sec:appendix} shows details of this digital data divide.
English-Hindi human-translated data is readily available for the domains we considered in this work. We obtained Sanskrit-based sub-word tokens from the available Hindi data for over 76\% of training and test instances. Furthermore, a word in one language may have several different translations in another language, depending on the context of usage. Providing the augmented input helps disambiguate the domain of the word. See Section~\ref{sec:posthoc} for a detailed analysis supporting this argument.

\section{Experiments and Results}
\label{sec:exp}

For the technical term translation task, we utilize the technical bilingual dictionary datasets provided by \citet{maheshwari2024lexgen} which is a dataset curated from CSTT \footnote{https://cstt.education.gov.in/en} dictionaries. The dataset consists of word-level translations from English to 6 Indian languages across 3 domains, \emph{viz.,} administrative, biotechnology, and chemistry, and has 9094 terms in the training data and 1285 in the test data for all domains combined. We obtained Sanskrit-based inputs for all data instances by applying our approach of generating Sanskrit-based additional inputs. We use chrF++ \cite{popovic2017chrf++} as the evaluation metric for all the experiments under this task.

\subsection{Experiments on the SWS Task}
For the SWS task, we use word-level accuracy as the evaluation metric. To compare against \cite{sandhan2022translist}, we also calculate sentence level perfect match (PM) for SIGHUM and hackathon datasets. We utilize the pre-trained checkpoint of the base variant of the ByT5 model available via Huggingface \footnote{https://huggingface.co/google/byt5-base} and fine-tune it over the UoH+SandhiKosh, SIGHUM dataset, and hackathon datasets as three separate experiments. For details about choice of ByT5 and Experiments with SWS Task (see Appendix \ref{appendix:sws}).

\begin{table*}[t!]
\centering
\resizebox{.8\textwidth}{!}{%
\begin{tabular}{@{}ccccccccc@{}}
\toprule
Test Dataset & Model & Hindi & Marathi & Gujarati & Kannada & Tamil & Odia & Average \\ \midrule

\multirow{2}{*}{Administrative}
& NLLB & 50.23 & 45.42 & 43.35 & 45.68 & \textbf{44.13} & 43.22 & 45.33 \\
 & NLLB + Sanskrit & \textbf{54.74} & \textbf{46.07} & \textbf{45.82} & \textbf{47.25} & 44.07 & \textbf{44.45} & \textbf{47.07} \\ \midrule
\multirow{2}{*}{Biotechnology} 
& NLLB & 53.52 & 51.91 & 3.79 & 12.38 & 18.46 & 17.16 & 26.20 \\
 & NLLB + Sanskrit & \textbf{60.63} & \textbf{60.73} & \textbf{13.09} & \textbf{29.20} & \textbf{37.89} & \textbf{35.82} & \textbf{39.56} \\ \midrule
 \multirow{2}{*}{Chemistry} 
& NLLB & 48.96 & 50.64 & 8.19 & 16.59 & 17.43 & 20.31 & 27.02 \\
 & NLLB + Sanskrit & \textbf{54.36} & \textbf{55.35} & \textbf{17.41} & \textbf{29.51} & \textbf{33.04} & \textbf{34.07} & \textbf{37.29} \\
\bottomrule
\end{tabular}
}
\caption{chrF++ scores on the administrative, biotechnology, and chemistry domains for models with and without additional Sanskrit-based input.}
\label{tab:exp1}
\end{table*}
%

%


%
\subsection{Experiments on the Technical Term Translation Task}
\label{sec:exp_trans}
For this task, we have two experimental settings, both formulated as text-to-text translation. In the first setting, we train and test the NMT model \textbf{NLLB} \cite{costa2022no} over all 6 language pairs across 3 domains. In the second setting, we train the model on Hindi, Gujarati, and Tamil across 3 domains and test it over Marathi, Kannada, and Odia across the same domains, which can be considered as a zero-shot setting. In the \textbf{baseline} configuration for this task, the model is fed with English input only. In the configuration corresponding to the proposed method, the English input is augmented with additional Sanskrit-based input prepared as discussed in Section~\ref{sec:sbad}. We utilize the pre-trained 1.3B parameter checkpoint of the NLLB model available via Huggingface\footnote{https://huggingface.co/facebook/nllb-200-1.3B} and fine-tune it over the technical domain dictionary data for both experimental settings. 

\paragraph{Results.} Table~\ref{tab:exp1} reports the comparison of chrF++ scores obtained by finetuning the NMT model with English-only input (NLLB) and with augmented input (NLLB+Sanskrit) under the first experimental setting. In Section \ref{sec:zeroshot} we analyze the performance of the model with and without additional input in a zero-shot setting. Across experiments, there's a consistent performance gain with the lexically informed input. Our method archives an average improvement of \textbf{8.46} chrF++ scores.
We also provide a detailed post-hoc analysis of the predictions in Section~\ref{sec:posthoc}

\subsubsection*{Zero-Shot Translation}
\label{sec:zeroshot}
Table~\ref{tab:exp3} shows the performance of the translation model without Sanskrit input (NLLB) and with Sanskrit input (NLLB+Sanskrit) when trained on Hindi, Gujarati, and Tamil, and evaluated on Marathi, Kannada, and Odia across 3 domains \emph{viz.,} Administration, Biotechnology, and Chemistry. Performance in terms of chrF++ scores shows that the translation with the Sanskrit augmented input consistently provides better translations as compared to the English-only input across different languages and domains. This proves the efficacy of Sanskrit-based additional input for capturing multilingual nuances.
\begin{table}[h]
\centering
\resizebox{0.5\textwidth}{!}{
\begin{tabular}{@{}cccccc@{}}
\toprule
Test Dataset & Model & Marathi & Kannada & Odia & Average\\ \midrule
\multirow{2}{*}{Administrative}
 & NLLB & 41.42 & 44.03 & 40.57 & 42.01 \\
 & NLLB + Sanskrit & \textbf{43.26} & \textbf{45.71} & \textbf{42.02} & \textbf{43.66} \\ \midrule
\multirow{2}{*}{Biotechnology } 
 & NLLB & 44.42 & 27.83 & 29.37 & 33.87 \\
 & NLLB + Sanskrit & \textbf{53.79} & \textbf{40.32} & \textbf{37.76} & \textbf{43.96} \\ \midrule
 \multirow{2}{*}{Chemistry} 
 & NLLB & 41.62 & 28.41 & 26.99 & 32.34 \\
 & NLLB + Sanskrit & \textbf{49.71} & \textbf{39.11} & \textbf{34.13} & \textbf{40.98} \\
 \bottomrule
\end{tabular}
}
\caption{chrF++ scores on administrative, biotechnology, and chemistry for unseen languages, namely, Kannada, Marathi, and Odia for zero-shot setting.}
\label{tab:exp3}
\end{table}

\subsection{Post-hoc analysis}
\label{sec:posthoc}
In this section, we present our detailed analysis of a subset of the results of the lexicon translation task. Unlike a regular translation task, which includes a complete sentence and paragraphs, we deal with a single word or phrase here. Such a short input may have many different possible translations in the target language, either the translations that can be used interchangeably or those that may be varied with the context of its usage. The evaluation metrics like BLEU and chrF may not effectively capture the quality of translation as it is obtained by comparison of the predictions with the available ground truth data. The ground truth data may have a single or limited number of meaningful translations, and as a result, a different but correct prediction may be penalised. \\
We followed the Human Post-hoc evaluation as per \citet{maheshwari2024lexgen} for the same two additional languages as presented by them \emph{viz.,} Punjabi and Malayalam, using the same subset of input data and metrics. Our goal is to understand the practical utility of the generated lexicon in the respective languages and the extent to which they may be helpful in translating technical books from English to low-resource Indian languages. We achieved an R@1 score of 0.53 and 0.46 for Punjabi and Malayalam, respectively, compared to 0.51 and 0.38 scores obtained by LexGen. The R@3 score for Malayalam is 0.72, comparable to 0.71 for LexGen, while the score for Punjabi was slightly lower, at 0.92, compared to 0.95 for LexGen.
We also present detailed analysis of the translation results by a comparative study of the outputs in both the input settings, i.e., with and without the Sanskrit-based augmented output (See Appendix~\ref{appendix:posthoc}).

\section{Conclusion}
In this work, we addressed the task of Sanskrit Word Segmentation (SWS) with a character-level Transformer model, achieving superior segmentation performance on two benchmark datasets and competitive performance on another benchmark dataset.. Furthermore, we propose to leverage the significant vocabulary overlap among Indian languages, utilizing data from the relatively resource-rich Hindi language which highlights the potential of cross-linguistic resource sharing to boost performance in low-resource language tasks.


\section*{Limitations}

To generate Sanskrit-based input, we rely on the available Hindi data. Though the availability of Hindi resources is much higher than that of other Indian languages, its digital data richness is considerably lower than that of English. \\
Not all languages exhibit significant vocabulary overlap with Sanskrit, and in such cases, our proposed method may have limited applicability for lexicon generation.

\section*{Acknowledgments}
We acknowledge IIT Bombay and BharatGen for providing resources and support for the project. Author Karthika acknowledges the PhD fellowship grant from the TCS Research Foundation. We also extend our appreciation to the reviewers for their valuable feedback.

\bibliography{custom,anthology}

\begin{thebibliography}{22}
\providecommand{\natexlab}[1]{#1}

\bibitem[{Aralikatte et~al.(2018)Aralikatte, Gantayat, Panwar, Sankaran, and Mani}]{aralikatte2018sanskrit}
Rahul Aralikatte, Neelamadhav Gantayat, Naveen Panwar, Anush Sankaran, and Senthil Mani. 2018.
\newblock Sanskrit sandhi splitting using seq2 (seq)\^{} 2.
\newblock \emph{arXiv preprint arXiv:1801.00428}.

\bibitem[{Bhardwaj et~al.(2018)Bhardwaj, Gantayat, Chaturvedi, Garg, and Agarwal}]{bhardwaj2018sandhikosh}
Shubham Bhardwaj, Neelamadhav Gantayat, Nikhil Chaturvedi, Rahul Garg, and Sumeet Agarwal. 2018.
\newblock Sandhikosh: A benchmark corpus for evaluating sanskrit sandhi tools.
\newblock In \emph{Proceedings of the Eleventh International Conference on Language Resources and Evaluation (LREC 2018)}.

\bibitem[{Costa-juss{\`a} et~al.(2022)Costa-juss{\`a}, Cross, {\c{C}}elebi, Elbayad, Heafield, Heffernan, Kalbassi, Lam, Licht, Maillard et~al.}]{costa2022no}
Marta~R Costa-juss{\`a}, James Cross, Onur {\c{C}}elebi, Maha Elbayad, Kenneth Heafield, Kevin Heffernan, Elahe Kalbassi, Janice Lam, Daniel Licht, Jean Maillard, et~al. 2022.
\newblock No language left behind: Scaling human-centered machine translation.
\newblock \emph{arXiv preprint arXiv:2207.04672}.

\bibitem[{Dave et~al.(2021)Dave, Singh, AP, and Lall}]{dave2021neural}
Sushant Dave, Arun~Kumar Singh, Dr~Prathosh AP, and Prof~Brejesh Lall. 2021.
\newblock Neural compound-word (sandhi) generation and splitting in sanskrit language.
\newblock In \emph{Proceedings of the 3rd ACM India Joint International Conference on Data Science \& Management of Data (8th ACM IKDD CODS \& 26th COMAD)}, pages 171--177.

\bibitem[{G{\'e}rard(2003)}]{gerard2003lexicon}
Huet G{\'e}rard. 2003.
\newblock Lexicon-directed segmentation and tagging of sanskrit.
\newblock In \emph{XIIth World Sanskrit Conference, Helsinki, Finland, Aug}, pages 307--325. Citeseer.

\bibitem[{Goyal and Huet(2013)}]{goyal2013completeness}
Pawan Goyal and G{\'e}rard Huet. 2013.
\newblock Completeness analysis of a sanskrit reader.
\newblock In \emph{Proceedings, 5th International Symposium on Sanskrit Computational Linguistics. DK Printworld (P) Ltd}, pages 130--171. Citeseer.

\bibitem[{Hellwig(2010)}]{hellwig2010dcs}
Oliver Hellwig. 2010.
\newblock Dcs-the digital corpus of sanskrit. heidelberg (2010-2021).
\newblock \emph{URL http://www.sanskritlinguistics.org/dcs/index.php.}

\bibitem[{Hellwig and Nehrdich(2018)}]{hellwig-nehrdich-2018-sanskrit}
Oliver Hellwig and Sebastian Nehrdich. 2018.
\newblock \href {https://doi.org/10.18653/v1/D18-1295} {{S}anskrit word segmentation using character-level recurrent and convolutional neural networks}.
\newblock In \emph{Proceedings of the 2018 Conference on Empirical Methods in Natural Language Processing}, pages 2754--2763, Brussels, Belgium. Association for Computational Linguistics.

\bibitem[{Hudelson(1987)}]{hudelson1987role}
Sarah Hudelson. 1987.
\newblock The role of native language literacy in the education of language minority children.
\newblock \emph{Language Arts}, 64(8):827--841.

\bibitem[{Huet(2003)}]{huet2003towards}
G{\'e}rard Huet. 2003.
\newblock Towards computational processing of sanskrit.
\newblock In \emph{International Conference on Natural Language Processing (ICON)}, pages 40--48.

\bibitem[{Krishna et~al.(2017)Krishna, Satuluri, and Goyal}]{krishna2017dataset}
Amrith Krishna, Pavankumar Satuluri, and Pawan Goyal. 2017.
\newblock A dataset for sanskrit word segmentation.
\newblock In \emph{Proceedings of the Joint SIGHUM Workshop on Computational Linguistics for Cultural Heritage, Social Sciences, Humanities and Literature}, pages 105--114.

\bibitem[{Krishnan et~al.(2020)Krishnan, Kulkarni, and Huet}]{krishnan2020validation}
Sriram Krishnan, Amba Kulkarni, and G{\'e}rard Huet. 2020.
\newblock Validation and normalization of dcs corpus using sanskrit heritage tools to build a tagged gold corpus.
\newblock \emph{arXiv preprint arXiv:2005.06545}.

\bibitem[{Kumar et~al.(2010)Kumar, Mittal, and Kulkarni}]{kumar2010sanskrit}
Anil Kumar, Vipul Mittal, and Amba Kulkarni. 2010.
\newblock Sanskrit compound processor.
\newblock In \emph{Sanskrit Computational Linguistics: 4th International Symposium, New Delhi, India, December 10-12, 2010. Proceedings}, pages 57--69. Springer.

\bibitem[{Kunchukuttan and Bhattacharyya(2016)}]{kunchukuttan2016faster}
Anoop Kunchukuttan and Pushpak Bhattacharyya. 2016.
\newblock Faster decoding for subword level phrase-based smt between related languages.
\newblock \emph{arXiv preprint arXiv:1611.00354}.

\bibitem[{Maheshwari et~al.(2024)Maheshwari, Singh, NJ, Bhatt, Jyothi, and Ramakrishnan}]{maheshwari2024lexgen}
Ayush Maheshwari, Atul~Kumar Singh, Karthika NJ, Krishnakant Bhatt, Preethi Jyothi, and Ganesh Ramakrishnan. 2024.
\newblock Lexgen: Domain-aware multilingual lexicon generation.
\newblock \emph{arXiv preprint arXiv:2405.11200}.

\bibitem[{Nehrdich et~al.(2024)Nehrdich, Hellwig, and Keutzer}]{nehrdich-etal-2024-one}
Sebastian Nehrdich, Oliver Hellwig, and Kurt Keutzer. 2024.
\newblock \href {https://doi.org/10.18653/v1/2024.findings-emnlp.805} {One model is all you need: {B}y{T}5-{S}anskrit, a unified model for {S}anskrit {NLP} tasks}.
\newblock In \emph{Findings of the Association for Computational Linguistics: EMNLP 2024}, pages 13742--13751, Miami, Florida, USA. Association for Computational Linguistics.

\bibitem[{Popovi{\'c}(2017)}]{popovic2017chrf++}
Maja Popovi{\'c}. 2017.
\newblock chrf++: words helping character n-grams.
\newblock In \emph{Proceedings of the second conference on machine translation}, pages 612--618.

\bibitem[{Raffel et~al.(2020)Raffel, Shazeer, Roberts, Lee, Narang, Matena, Zhou, Li, and Liu}]{raffel2020exploring}
Colin Raffel, Noam Shazeer, Adam Roberts, Katherine Lee, Sharan Narang, Michael Matena, Yanqi Zhou, Wei Li, and Peter~J Liu. 2020.
\newblock Exploring the limits of transfer learning with a unified text-to-text transformer.
\newblock \emph{Journal of machine learning research}, 21(140):1--67.

\bibitem[{Sachin(2007)}]{sachin2007sandhi}
Kumar Sachin. 2007.
\newblock Sandhi splitter and analyzer for sanskrit (with reference to ac sandhi).
\newblock \emph{Mphil degree at SCSS, JNU (submitted, 2007)}.

\bibitem[{Sandhan et~al.(2022)Sandhan, Singha, Rao, Samanta, Behera, and Goyal}]{sandhan2022translist}
Jivnesh Sandhan, Rathin Singha, Narein Rao, Suvendu Samanta, Laxmidhar Behera, and Pawan Goyal. 2022.
\newblock Translist: A transformer-based linguistically informed sanskrit tokenizer.
\newblock \emph{arXiv preprint arXiv:2210.11753}.

\bibitem[{Sriram et~al.(2023)Sriram, Kulkarni, and Huet}]{sriram-etal-2023-validation}
Krishnan Sriram, Amba Kulkarni, and G{\'e}rard Huet. 2023.
\newblock \href {https://aclanthology.org/2023.wsc-csdh.3} {Validation and normalization of {DCS} corpus and development of the {S}anskrit heritage engine{'}s segmenter}.
\newblock In \emph{Proceedings of the Computational {S}anskrit {\&} Digital Humanities: Selected papers presented at the 18th World {S}anskrit Conference}, pages 38--58, Canberra, Australia (Online mode). Association for Computational Linguistics.

\bibitem[{Xue et~al.(2022)Xue, Barua, Constant, Al-Rfou, Narang, Kale, Roberts, and Raffel}]{xue2022byt5}
Linting Xue, Aditya Barua, Noah Constant, Rami Al-Rfou, Sharan Narang, Mihir Kale, Adam Roberts, and Colin Raffel. 2022.
\newblock Byt5: Towards a token-free future with pre-trained byte-to-byte models.
\newblock \emph{Transactions of the Association for Computational Linguistics}, 10:291--306.

\end{thebibliography}

\clearpage
\appendix

\section{Appendix}
\label{sec:appendix}
\subsection{Normalisation}
\label{sec:norm}
anusvāra (ṃ), is a symbol used in all Indian language scripts to denote a type of nasal sound. According to Sanskrit grammatical rules, when this symbol precedes one of the first 4 characters in each of the consonant group called vargās (ka/ca/ṭa/ta/pa), it needs to be converted to the respective fifth characters (pañcamākṣara) of the vargās (ṅ/ñ/ṇ/n/m). This rule may not be followed in other Indian languages. Since our sub-word segmentation model is trained on Sanskrit, and applied on Hindi data for the translation task, we normalise all the data by converting all occurences of anusvāra to the corresponding pañcamākṣara, before passing it to our model for segmentation.\\

\subsection{sandhi}
Sanskrit and other Indian languages have common usage of compound words, which are formed from multiple subwords. When two words are combined, the language expects certain rules to be followed at the word boundaries. Such a change in the word boundary forming a compound word, is termed as sandhi (the word has a meaning of \textit{junction}. In Sanskrit, there are specific rules for the joining of subwords to form a compound, depending on the ending character of the first and the beginning character of the second word. We specify these rules as the \textit{sandhi rules} in this paper. Similarly, splitting of the sandhi will also need to follow the reverse process, which is not as straightforward as sub-word joining. In the paper, we specify the process of sandhi splitting as \textit{sandhivicchēda}. Following are some examples of sandhivicchēda (1) tatrāpi = tatra + api; (2)narēndra = nara + indraḥ

\subsection{Post-hoc Analysis of Generated Technical Lexicons}
\label{appendix:posthoc}
Table~\ref{tab:post_hoc} shows some qualitative, post hoc analysis of the prediction results. The analysis shows that the augmented input 
\begin{asparaitem} 
\item Assists the model to disambiguate between multiple possible outputs (synonyms) and obtain the contextually apt term.
\begin{itemize}
    \item Examples 1 and 2 in table~\ref{tab:post_hoc} are from the Administration domain, with Kannada as the required target language. The translations generated by the model with only the English input are meaningful but in different contexts. The word \textit{mass} is considered by the model, in the meaning of \textit{the amount of matter in an object }, while the expected meaning is mass as used in \textit{population }
    \item Similarly, the word \textit{composition} is expected to take the meaning of \textit{composing music or poetry}, while the meaning taken by the model is \textit{the process of combining parts of something to whole}. Example 4 shows a similar trend in Marathi in Biotechnology domain.\\
For the above examples, our model is able to disambiguate the intended meaning and generate the expected output.
\end{itemize}
    \item Examples 3 is a sample where the output generated with English-only input is incorrect, while the augmented input generates correct output. 
\end{asparaitem}

\begin{table*}[h]
\centering
\resizebox{\textwidth}{!}{%
\begin{tabular}{l@{}ccccc@{}}
\toprule
 & Technical term (English) &  Domain; Language&  Augmented input \footnote{We provide the augmented input in the target language script. Here we add ISO-15919 transliteration for better readability}& \multicolumn{2}{c}{Prediction with}\\
          &&  &  &  English only input& Sanskrit-based augmented input\\ \hline
          1&mass&  Administration; Kannada&  mass <SEP> jana <isep> samūha&  dravyamāna& jana-samūha\\
  2&composition& Administration; Kannada& composition <SEP> racanā& samyōjane&racanā\\ 
  3&brood& Biotechnology; Marathi& brood <SEP> bhrūṇa& prajanana & bhrūṇa\\
  4&transformation& Biotechnology; Marathi& transformation <SEP> rūpa <isep> antaraṇa & parivartana & rūpāntara\\
  5&injection& Biotechnology; Marathi& injection <SEP> antaḥ <isep> kṣēpaṇa &injēkśana &antaḥ-kṣēpaṇa\\
 
\bottomrule
\end{tabular}
}

\caption{Post hoc Qualitative Analysis of Technical term translation results}
\label{tab:post_hoc}
\end{table*}

We notice that, the performance difference with and without augmented input is less in the administrative domain when compared to other domains. With the observations from the predictions, we arrive at the following reasonings.
The words in this domain are very frequently used by people in all languages. The model predictions with augmented input results in many archaic words, which are currently not in use, or the usage is highly infrequent. A word can have a large number of synonyms, and the number of words in the reference list of the ground truth, is limited, which mostly do not include the archaic words. Because of these reasons, we do not see a large jump in the performance with augmented input in this domain. This observation is especially true with languages like Tamil, in which there is a significant number of non-Sanskrit originated words, which may be more commonly in use.
In both experimental settings, we observe that the gain is more in case of the biotechnology and chemistry domains as compared to the administrative domain. This behavior can be attributed to the pre-training of the NLLB model on massive generic domain data which has considerable overlap with the administrative domain data.

\section{Sanskrit Word Segmentation Task}
\subsection{Data - SWS Task}

For the SWS task, following \citet{dave2021neural} and \citet{sandhan2022translist}, we use three publicly available benchmark datasets, \textit{UoH corpus}\footnote{https://sanskrit.uohyd.ac.in/Corpus/} combined with the \textit{SandhiKosh dataset} \cite{bhardwaj2018sandhikosh}, \textit{SIGHUM dataset} \cite{krishna2017dataset}, and \textit{hackathon dataset} \cite{krishnan2020validation}. These datasets are carefully curated subsets of a larger corpus DCS \cite{hellwig2010dcs}. The UoH corpus+SandhiKosh dataset has 62273 and 15569 instances as train and test sets. For this dataset, we apply the pruning technique mentioned in \cite{dave2021neural} to filter out invalid instances. The size of the training, validation, and test sets for the SIGHUM dataset are 97000, 3000, and 4200, respectively, and for the hackathon dataset, it is 90000, 10332, and 9963, respectively. Contemporary deep-learning methodologies have demonstrated enhanced performance when utilizing the SLP1 script for Sanskrit. Consequently, we have prepared all datasets in the SLP1 script to leverage these performance improvements. 

\begin{table}[h]
\centering
\resizebox{0.45\textwidth}{!}{%
\begin{tabular}{@{}p{3cm}p{3cm}p{1cm}@{}}
\toprule
\textbf{Model} & \textbf{LPA} & \textbf{SPA} \\ \midrule
    JNU & - & 8.1 \\
    UoH & - & 47.2 \\
    INRIA & - & 59.9 \\
    DD-RNN & 95.0 & 79.5 \\
    Sandhi Prakarana & 92.3 & 86.8 \\
    ByT5& \textbf{97.2} & \textbf{93.5} \\
    \bottomrule
\end{tabular}
}
\caption{Location prediction accuracies (LPA) and split prediction accuracies (SPA) for different methods on the UoH+SandhiKosh dataset. }
\label{tab:byt5onsandhikosh}
\end{table}
\subsection{Experiment Details}
\label{appendix:sws}
\paragraph{Baselines.} For the experiments performed over the \textit{UoH+SandhiKosh} dataset, we compare our method against \textbf{Sandhi Prakarana} \cite{dave2021neural}, \textbf{DD-RNN}  \cite{aralikatte2018sanskrit}, and  3 sandhi spitter tools viz (i) \textit{JNU Splitter} \cite{sachin2007sandhi}, (ii) \textit{UoH Splitter} \cite{kumar2010sanskrit}, and (iii) \textit{INRIA Sanskrit Heritage Reader} \cite{huet2003towards,goyal2013completeness}. We reproduce and report the scores reported by \citet{dave2021neural}. For DD-RNN and the 3 sandhi tools, we report the scores reported in \cite{aralikatte2018sanskrit} and \cite{dave2021neural}. For the experiments performed over the \textit{SIGHUM} and \textit{hackathon} datasets, we compare our method against \textbf{TransLIST} \cite{sandhan2022translist} and \textbf{rcNN-SS} \cite{hellwig-nehrdich-2018-sanskrit}.

\paragraph{Results.} Tables~\ref{tab:byt5onsandhikosh} and~\ref{tab:byt5onsighumandhackathon} report the performance of our methodology compared with the baselines over the respective datasets. Table~\ref{tab:byt5onsandhikosh} shows that our methodology outperforms all other baselines in terms of both Location Prediction Accuracy (LPA) and Split Prediction Accuracy (SPA) with absolute gains of \textbf{\textit{4.86}} and \textbf{\textit{6.72}}, respectively, on the \textit{UoH+SandhiKosh} dataset. 
TransLIST \citet{sandhan2022translist} utilizes a set of potential split candidates from SHR (referred to as LIST in their paper), which provides additional linguistic information for segmentation. Our model is not linguistically informed like this as we feed only the compound word to the model. Hence, our method is not strictly comparable with the results shown in row 2 of Table~\ref{tab:byt5onsighumandhackathon}. Nevertheless, our method outperforms all other models on three out of four evaluation metrics when tested on \textit{hackathon} dataset. On \textit{SIGHUM} dataset, our method achieves competitive scores. \citet{sandhan2022translist} also reported the performance of their model without the LIST module, as shown in row 3 (\underline{TransLIST}). The model without the LIST step is more comparable to our setting and we outperform this result as well, while failing to outperform the scores in row 2. As a separate experiment, we provide SHR input to our model for \textit{SIGHUM} data which outperforms TransLIST on PM metric achieving a PM score of \textbf{94.31}.

\begin{table}[h]
\begin{center}
    \resizebox{.5\textwidth}{!}{%
    \begin{tabular}{l|cccc|cccc}
    \toprule
    Model & \multicolumn{4}{c|}{SIGHUM} & \multicolumn{4}{c}{Hackathon} \\ \cmidrule{2-9} 
     & P & R & F & PM & P & R & F & PM \\ \midrule
    rcNN-SS & 96.86 & 96.83 & 96.84 & 87.08 & 96.40 & 95.15 & 95.77 & 77.62 \\ \midrule
    TransLIST & \textbf{98.80} & \textbf{98.93} & \textbf{98.86} & \textbf{93.97} & \textbf{97.78} & 97.44 & 97.61 & 85.47 \\ \midrule
    \underline{TransLIST} & - & - & - & 86.10 & - & - & - & - \\ \midrule
    ByT5& 98.68 & 98.42 & 98.53 & 93.78 & 97.58 & \textbf{97.71} & \textbf{97.63} & \textbf{87.7} \\ \bottomrule
    \end{tabular}
    }
    \caption{Word-level Precision, Recall, F1 and sentence-level Perfect Match (PM) scores  on SIGHUM and hackathon.}
    \label{tab:byt5onsighumandhackathon}
\end{center}
\end{table}

\end{document}